\documentclass[%
 reprint,
 amsmath,amssymb,
 aps,
]{revtex4-2}

\usepackage{soul}
\usepackage{xcolor}
\usepackage{graphicx}
\usepackage{dcolumn}
\usepackage{bm}

\begin{document}

\preprint{APS/123-QED}

\title{Deterministic versus stochastic dynamical classifiers: opposing random adversarial attacks with noise}

\author{Lorenzo Chicchi}
\email{lorenzo.chicchi@unifi.it}
\affiliation{Department of Physics and Astronomy and INFN, University of Florence, Via Sansone 1 Sesto Fiorentino 50019, Italy}
\author{Duccio Fanelli}
\affiliation{Department of Physics and Astronomy and INFN, University of Florence, Via Sansone 1 Sesto Fiorentino 50019, Italy}
\author{Diego Febbe}
\affiliation{Department of Physics and Astronomy and INFN, University of Florence, Via Sansone 1 Sesto Fiorentino 50019, Italy}
\author{Lorenzo Buffoni}
\affiliation{Department of Physics and Astronomy and INFN, University of Florence, Via Sansone 1 Sesto Fiorentino 50019, Italy}
\author{Francesca Di Patti}
\affiliation{Department of Mathematics and Computer Science , University of Perugia, Via Vanvitelli 1, and \\ INFN, Sezione di Perugia, via A. Pascoli 23c, Perugia 06123, Italy}
\author{Lorenzo Giambagli}
\affiliation{Department of Physics and Astronomy and INFN, University of Florence, Via Sansone 1 Sesto Fiorentino 50019, Italy}
\author{Raffaele Marino}
\affiliation{Department of Physics and Astronomy and INFN, University of Florence, Via Sansone 1 Sesto Fiorentino 50019, Italy}





\begin{abstract}
The \textit{Continuous-Variable Firing Rate} (CVFR) model, widely used in neuroscience to describe the intertangled dynamics of excitatory biological neurons, is here trained and tested as a veritable dynamically assisted classifier. To this end the model is supplied with a set of planted attractors which are self-consistently embedded in the inter-nodes coupling matrix, via its spectral decomposition. Learning to classify amounts to sculp the basin of attraction of the imposed equilibria, directing different items towards the corresponding destination target, which reflects the class of respective pertinence. A stochastic variant of the CVFR model is also studied and found to be robust to aversarial random attacks, which corrupt the items to be classified. This remarkable finding is one of the very many surprising effects which arise when noise and dynamical attributes are made to mutually resonate.  
\end{abstract}

\maketitle

\section{\label{sec:level1}Introduction}

Deep Neural Networks (DNNs) \cite{deng2014deep,goodfellow2016deep,lecun2015deep} are the state of the art for modern regression and classification problems in numerous fields \cite{unke2021machine, chicchi2023frontiers, biancalani2021deep, baldi2014searching,marino2023solving}. Their exceptional ability in recognizing patterns and resolving hidden correlations from data places DNNs at the forefront of the most sophisticated artificial intelligence tools  \cite{vaswani2017attention, chowdhary2020natural}. Despite being endowed with indisputable practical effectiveness, many aspects of DNNs remain still unclear and call for further foundational analysis. Searching for novel artificial neural network paradigms proves therefore important for different reasons: on the one side, new NN models and architectures could eventually enhance over current performance when challenged against specific  domains of application; on the other, innovative approaches - and associated formalism - could craft original, so far unexplored, scenarios to grasp the intimate essence of the most elusive aspects of DNNs functioning.

Among the existing classes of neural network models, recurrent neural networks (RNNs) \cite{hochreiter1997long, goldberg2022neural} are certainly noteworthy. RNNs have been successfully employed in several fields of research, demonstrating their irrefutable effectiveness in the widespread realm of time series analysis \cite{connor1994recurrent, hewamalage2021recurrent} and natural language processing (NLP) \cite{goldberg2022neural, wang2015learning, yao2018improved}. The distinctive characteristic of RNN models is the presence of an internal state vector, that is iteratively processed along with the supplied input vector. More specifically, RNNs (i) generate an output vector and (ii) simultaneously update the internal state vector. Working in this framework, the imposed depth (i.e., the number of layers that compose  sensible computing DNN architectures) mirrors the number of implemented iterations, veritable repetitions of the very same operations as stemming from an identical set of adjustable parameters. For what here relevant, with the RNNs, the concept of temporal evolution, although confined to the discrete time domain, earned hence the stage of automated machine learning.

Further, in \cite{chen2018neural}, a new class of neural networks called Neural Ordinary Differential Equations (nODE) was proposed, in which the temporal derivative of the internal state vector of a recurrent neural network is parameterized. Working with the temporal derivative is conceptually equivalent to taking the continuous limit of a RNNs. In fact, a scheme somehow equivalent to that proposed in \cite{chen2018neural}, with the explicit inclusion of a time decaying exponential term, was pioneered several years before and made popular under the name of Continuous Recurrent Neural Networks (CRNNs) \cite{funahashi1993approximation, beer1995dynamics}. Algorithms spanning the nODEs typology proved very effective to handle data that extend on the time domain. A successful example is provided by a recent evolution of nODEs, called Liquid Time Constant Networks (LTCs). This latter showed an unexpected ability in tackling complex self-driving tasks by just involving a modest number of computing nodes \cite{hasani2021liquid}.\\

\indent From a general standpoint, nODEs represent the first genuine example of how dynamical systems could be efficaciously used to solve regression and classification problems, with a displayed success score comparable to that reported for traditional DNNs. Moreover, having drawn an ideal bridge with the renewed field of continuous dynamical systems
makes it possible, at least in principle, to leverage on a vast arsenal of techniques, developed and tested for different purposes, to cast novel light onto DNNs, in the aim of removing, at least partially,  their wrapping curtain of opacity and mystery.  \\


\indent A variant of nODEs targeted to classification problems, and called SA-nODEs, has been recently proposed in \cite{marino2023SANODE}. The  model is made of $N$ linearly coupled nodes, each bearing a continuous state variable which explores the landscape of a locally imposed double well potential. As opposed to its original counterpart, the untrained SA-nODE model is a priori constructed so as to accommodate for a set of pre-assigned stationary stable attractors. This is achieved by resorting to an ad hoc spectral decomposition of the inter-nodes coupling matrix \cite{giambagli2021machine,chicchi2021training,chicchi2023recurrent, chicchi2023complex, giambagli2024student, buffoni2022spectral, chicchi2024automatic}. The entire evaluation procedure as carried out by SA-nODEs, from the input reading to the final classification output, unfolds within a dynamical process, accessible, before and after training, for comparative analysis. In particular, the training amounts to reshaping the basins of attraction of the target destinations by ensuring that each initial condition (i.e. the input vector) evolves towards its corresponding equilibrium. In this work we elaborate further on the SA-nODE perspective \cite{marino2023SANODE}, and report on manifold non trivial extensions of the method in regards with its original conception. 

First of all, we will here operate with the celebrated \textit{Continuous-Variable Firing Rate} (CVFR) model \cite{kim2019simple}, widely employed in computational neuroscience to reproduce in silico  the orchestrated dynamics of a population of mutually entangled biological neurons. At variance with the setting discussed in \cite{marino2023SANODE}, and as we shall later on clarify, the local reaction dynamics of the CVFR model are linear. Binary inter-nodes couplings are instead modulated by a non linear filter. Following SA-nODE approach, we have here devised a novel procedure, tailored for the problem at hand, to equip the CVFR with a set of stable attractors that become the target of the learnable dynamics. The model is then reformulated in terms of a discrete map, to be deployed on a RNN architecture. This allows in turn to perform a straightforward optimization of the residual parameters for the system to react differently to distinct classes of processed items. More specifically,  input items belonging to a given category, and supplied to the dynamical model as an initial condition, will be directed towards a designed equilibrium, made stable by the training procedure. The procedure will be tested against different datasets with the sole aim of establishing a proof of principle for the adequacy of the generalized SA-nODE approach. As a byproduct of the analysis, we will show in fact that a model of biomimetic inspiration, suitable complemented with spectral attributes, can be trained to handle non trivial classification tasks, with an accuracy score comparable with state of the art DNN implementations. Incidentally, we also note that the employed CVFR model is  known in the literature as the continuous Hopfield model, a reference framework for associative memory studies.

In the second part of this work we will move on to consider a stochastic version of the CVFR (or continuous Hopfield) model. As we shall explain, the imposed noise term is multiplicative in nature, namely it is self-consistently modulated as a function of the co-evolving state variables. More specifically, it is designed to actively perturb the underlying deterministic dynamics out of equilibrium, while being progressively silenced when the system approaches the crafted asymptotic attractors. The presence of the noise component enhances the robustness of the trained model to random adversarial attacks, as we shall prove with a dedicated campaign of numerical tests.

The paper is organized as follows.  In Section \ref{sec:model}, the model is introduced, along with the mathematical details that define the  adopted strategy for planting the asymptotic attractors. We will then discuss the deployment of the model on a RNN, review the steps that pertain to the model training and report on the classification performance for the chosen datasets (a reservoir of images of stylized letters and MINST, the celebrated dataset of images with hand written digits). Then, in Section \ref{sec:stochastic formulation}, we will discuss the stochastic variant of the SA-nODE as applied to the CVFR (Hopfield) model and quantify the ability of the system to oppose adversarial attacks as a function of the inherent noisy component. Finally, in Section \ref{sec:conclusion}, we will sum up and illustrate the conclusions of this study.

\section{Setting the mathematical model: deterministic version}
\label{sec:model}

In this section, we introduce the deterministic version of the \textit{Continuous-Variable Firing Rate} (CVFR) model. We will also discuss how to force the presence of a set of asymptotic attractors by properly crafting the matrix of internodes connections via the associated spectral attributes. Finally we shall also provide the relevant information to assess, ex post, the stability of the planted equilibria. 

\subsection{The CVFR (or continuous Hopfield) model}
\label{sec:firing rate model}

The dynamical system that we shall here train as an automatic classifier, following the SA-nODE recipe, is widely used in computational neuroscience to reproduce in silico the evolution of a population made of intertwined spiking biological neurons \cite{kim2019simple}. Consider a set of $N$ neurons and assign to each individual unit $i=1,...,N$ the scalar variable $x_i$. This latter can be assumed to represent the synaptic current of neuron $i$. The system is defined by the following set of ordinary differential equations:

\begin{equation}
    \label{eq:model_definition}
    \dot x_i =-x_i +\frac{1}{\sqrt{N}} \sum_j^N A_{ij} f(x_j)
\end{equation}

where $f(\cdot)$ stands for a suitable non linear function and $A_{ij}$ are the entries of the weighted adjacency matrix $A$ that defines the architecture of the computing network. The normalization factor $1/\sqrt{N}$ is just a convenient scaling of the matrix elements, the indirect target of the training as we highlight below. In general, the non linear function $f(\cdot)$ is assumed to be sigmoidal in shape, so as to mimic the mechanism of neuronal activation. The image of the function can span different intervals depending on the specific range of definition assigned to the dynamical variable $x_i$. In this work, we will assume an  activation of the Hill type, by postulating $f(x_i)={x_i^2}/{\left(c+x_i^2 \right)}$ with $c \in \mathbb{R}^+$. This choice has the sole scope of allowing for a transparent description of the procedure that we have engineered for planting the attractors. The results can be straightforwardly extended to alternative model formulations that accommodate for distinct, though biologically plausible, non linear functions.  Before continuing it is worth remarking that the above Eq. (\ref{eq:model_definition}) are a special case of the general class of models that goes under the nODE acronym. In fact, it can be cast in the form $\dot x_i =g(x_i, t, \theta)$, where $g$ represents a neural network defined by parameters $\theta$.  In other words, the proposed model is a neural Ordinary Differential Equation with the time derivative of the internal state vector parameterized by a linear transformation, a nonlinearity, and an exponential decay term – a concept also found in continuous recurrent neural networks \cite{funahashi1993approximation}. As we will discuss, the learnable parameters are (a subset of) the elements of the adjacency matrix $A$ that we will hereafter assume to be defined via its spectral decomposition.

\subsection{How to impose beforehand crafted attractors for the CVFR dynamics}   
\label{sec:fixing_stationary_states}

We will show here how a spectral parameterization of the adjacency matrix $A$ can be invoked to enforce suitably engineered attractors of the collective neurons dynamics. 

To begin, let us assume that $\vec{x}^s$ identifies a stationary solution of Eq. (\ref{eq:model_definition}), namely that $\dot x^s_i=0$ $\forall$ $i$. Further suppose (and this will be enforced later) that $\psi_i={x^{s2}_{i}}/{\left(c+x^{s2}_{i}\right)}$ are the components of $\vec \psi$, an eigenvector of $A$ relative to eigenvalue $\lambda$. In other word, we postulate that $A \vec \psi = \lambda \vec \psi$. Then, Eq. (\ref{eq:model_definition}) readily yield the following self-consistent cubic equation:

\begin{equation}
\label{eq:stab_condition}
    x^s_i=\beta \lambda \frac{x^{s2}_{i}}{c+x^{s2}_{i}},
\end{equation}

where $\beta = 1/\sqrt{N}$ and which admits the solutions:

\begin{equation}
    \label{eq:solutions_pm}
x^s_i=0 \qquad  x^s_i = \frac{\beta \lambda \pm \sqrt{\beta^2 \lambda^2 - 4c}}{2}.
\end{equation}

Hereafter, we shall denote by $x_p$ and $x_m$ the two above non trivial solutions and assume $c<\frac{\lambda^2}{4N}$ to deal with real quantities. The analysis so far developed implies that any eigenvector $\vec \psi$ of $A$ with elements $\psi_i\in \left\{0, f(x_m), f(x_p)\right\}$, for $i=i,...,N$ and with associated eigenvalue $\lambda$, is a stationary solution of the system of ODEs (\ref{eq:model_definition}).
Stated differently, we have created an alphabet of three digits, $(0,x_m, x_p)$ which can be used at will to paint a virtually unlimited (bounded by $N$) collection of different stationary solutions $\vec{\psi}_k$ of the examined system. The request to be additionally met is that $\vec{\psi}_k$ are eigenvectors of matrix $A$ relative to the same eigenvalue $\lambda$, as we will assume in the following. More degrees of freedom can be accommodated for by removing the degeneracy on the spectrum. This amounts to assigning a distinct eigenvalue $\lambda_k$ to each $\vec{\psi}_k$, a choice that generates different alphabet triplets (except for the null entry) for crafting the eigenvectors.

As previously mentioned, assume ${A}=\Psi \Lambda \Psi^{-1}$, where $\Psi$ belongs to the set of real matrices ${R}^{N \times N}$ and $\Psi^{-1}$ stands for its inverse. Matrix $\Psi$ incorporates the eigenvectors of ${A}$, arranged as its columns; $\Lambda$, also in ${R}^{N \times N}$, is a diagonal matrix containing ${A}$'s eigenvalues. The choice  of decomposing the interaction matrix in reciprocal space echoes the spectral approach to machine learning, as outlined in references \cite{giambagli2021machine, buffoni2022spectral, chicchi2021training, chicchi2023recurrent}. To plant a priori $K$ non linear stable attractors (where $ K $ stands for the total number of classes to be eventually categorised), it suffices to force the first $K$ columns of  $\Psi$ to be identically equal to $\vec{\psi}_k$, with $k=1,...,K$, the eigenvectors generated according to the recipe discussed in the preceding paragraph. The corresponding eigenvalues, i.e. the first $K$ entries of $\Lambda$, will be set to the same value $\lambda$. This latter is the eigenvalue that ultimately enters in the definition of the aforementioned alphabet, for what concerns the non trivial elements $x_p$ and $x_m$. The unset $\left( N-K \right) \times N$ entries of $\Psi$, together with the remaining $N-K$ eigenvalues, defines the target of the training. Acting on this latter pool of trainable parameters, we will teach to the examined dynamical system how to steer towards different imposed equilibria, depending on the characteristic of the items, supplied as an input, to be eventually classified. It is worth stressing that, at variance of the original SA-nODE formulation, the planted attractors do not coincide with the eigenstate of the coupling matrix $A$, due to the non linear function that modulates the interaction term. Also, the stability of the planted attractors cannot be enforced a priori, at least with the Hill type of non linearity that we have here chosen for pedagogical reasons. Other non linear functions can be nonetheless selected that will make it possible for the stability of the imposed attractors to be set before training, as we will report in a separate contribution. In the following, we will discuss the conditions that should be met for the attractors to be linearly stable and use this knowledge to certify ex post that the stability of the target destinations has been achieved, as a byproduct of the training. 

\subsection{On the linear stability of the imposed attractors}  

Let us elaborate on the conditions that underlies linear stability of a given stationary solution $\vec{x}^s$ of system (\ref{eq:model_definition}). Recall that we are by definition interested in a particular class of attractors: the entries of $\vec{x}^s$ are just limited to the triplet $(0, x_m, x_p)$ and the non linear image of the stationary solution, via $f(\dot{})$, is an eigenvector of $A$ relative to 
eigenvalue $\lambda$. With these premises, we focus on the $i-th$ component of $\vec{x}^s$ and insert a small perturbation $\delta x_i$ as stipulated by:

\begin{equation}
    \label{eq:linear_approximation}
    x_i = x^s_{i}+\delta x_i.
\end{equation}

By introducing the latter expression in Eq. (\ref{eq:model_definition}) and expanding to the first order in $\delta x_i$ yields:

\begin{equation}
    \dot{\delta x_i}= -\delta x_i + \beta \sum_j^N  A_{ij}f'(x^s_{j})\delta x_j,
\label{eq:perturbation_linearization}
\end{equation}

where use has been made of the definition of stationary solution. By defining the diagonal matrix $\mathcal{F}'(x)$ with elements $[\mathcal{F}']_{ij}= f'(x^s_{j}) \delta_{ij}$ and the matrix $\mathbf{J} = -\mathcal{I} + \beta A \mathcal{F}'$, where $\mathcal{I}$ is the identity matrix,  Eq. \ref{eq:perturbation_linearization} can be rewritten as

\begin{equation}
    \dot{\delta \vec{x}}= \mathbf{J}\delta \vec{x}.
\end{equation}

To assess the linear stability of the stationary solution $\vec{x}^s$ it is therefore sufficient to numerically compute the eigenvalues of  matrix $\mathbf{J}$. If all eigenvalues display a negative defined real part, the stationary state is stable. As we will show, the training process steers the model towards a regime where the defined stationary equilibria are de facto stable.

\section{Training the Deterministic CVFR model to achieve classification}
\label{sec:classification_problem}

Following the analysis developed above, we are in a position to discuss  the training process that will transform the dynamical system into a veritable classification algorithm. We here recall that the target of the training are just the components of the matrix $\Psi$, which are not associated to the embedded eigenvectors, as well as the free eigenvalues, i.e. those that are not frozen to the value $\lambda$ and which can be therefore self-consistently learned. 

\begin{figure}[htbp]
    \includegraphics[width=0.4 \textwidth]{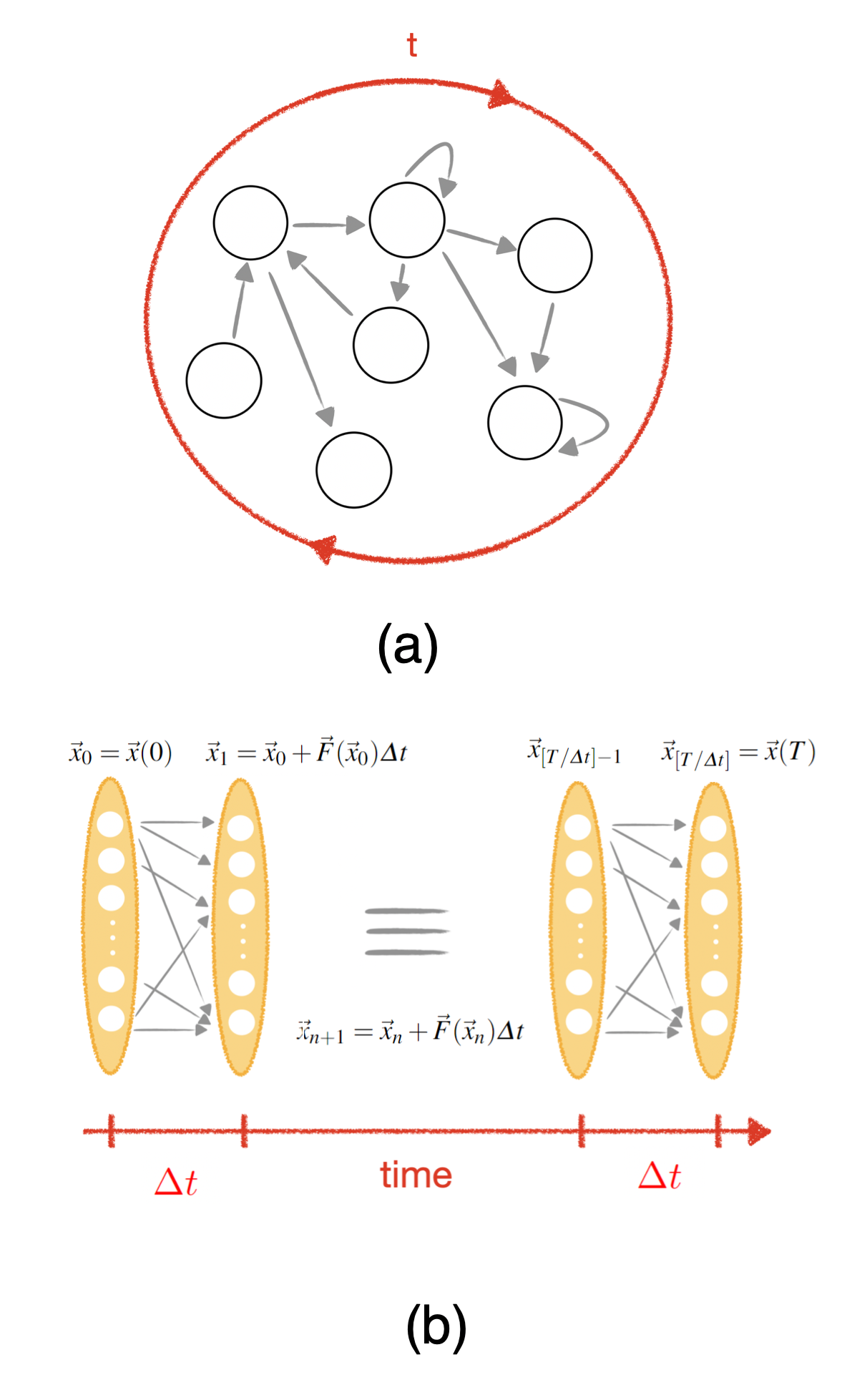} \\
    \caption{Panel (a): Schematic representation of the dynamical model employed. It is made by interacting neurons, each being subject to a linear local dynamics. The non linear activation term hides inside the arrow that points to the directed inter-nodes interactions. In the application that we shall discuss, each neuron is uniquely associated to a single pixel of the image to be analyzed. Panel (b): Schematic representation of the discrete Euler map deployed on a feedforward neural architecture.
    }
    \label{fig1NN}
\end{figure}

As the system evolves in time, the values of $\vec{x}$ get consequently updated, following the model prescription as dictated by Eq. (\ref{eq:model_definition}). In our scheme, the updating of the state variable is performed by an Euler algorithm, implemented as a recurrent neural network. To clarify the adopted algorithmic procedure, we display in Fig. \ref{fig1NN} a graphical portrait, adapted from \cite{marino2023SANODE}. Identical layers made of $N$ nodes are linked by a linear coupling matrix $\mathbf{A}$, defined by its spectral decomposition. Here, the parameters to be trained are eventually stored. Edges linking nodes implement a non linear filter, of the type discussed in the model setting. Conversely, local reactions taking place at each node/neuron yield an exponential decay. This is reflected in the linear term on the right hand side of Eq. (\ref{eq:model_definition}).  Time flows along the horizontal axis reported in Fig. \ref{fig1NN}. Nearby 
layers of the imposed feedforward deep architecture are separated by a finite amount in time, $\Delta t << 1$. By recasting in such terms the studied dynamical model allows us to address the sought optimization via standard numerical tools, as made available by the machine learning community. If $\Delta t$ is taken small enough, the trained discrete model will behave like its continuous analogue. Other integration methods, including Runge-Kutta, can be adopted \cite{marino2023SANODE} at the price of a complexification of the feedforward architecture on which the dynamical system is being deployed. 

We now proceed by introducing the two classification problems addressed in this work to test the adequacy of the proposed scheme. We chose to deal with image classification, in line with what was done with the original SA-nODE setting \cite{marino2023SANODE}.

The first dataset contains three classes of $7 \times 7$ images showing three distinct letters (A, B, C). Each image is deliberately corrupted by replacing the intensity of $20\%$ of the unperturbed pixels, with a random scalar drawn from a uniform distribution. The task's objective is to associate each noisy image with the correct class. The second employed dataset is the well-known MNIST dataset \cite{deng2012mnist}, which contains images of handwritten digits. The respective sizes of the system are $N= 7 \times 7 = 49$ (for the database made of letters) and $N=28 \times 28 = 784$ (for the standard version of MNIST).

In both cases, the image pixel intensity vectors, normalized to the range of 0 to 1, are used as an initial conditions for the evolution of the dynamical system defined by Eq. (\ref{eq:model_definition}). In practical terms, images are unwrapped as vectors of size $N$, and each pixel is mapped in one individual neuron of the systems' collection. The number of chosen attractors reflects the classes to be identified, $K=3$ and $K=10$ respectively.

\begin{figure}[h!]
    \centering
    \includegraphics[width=0.45\textwidth]{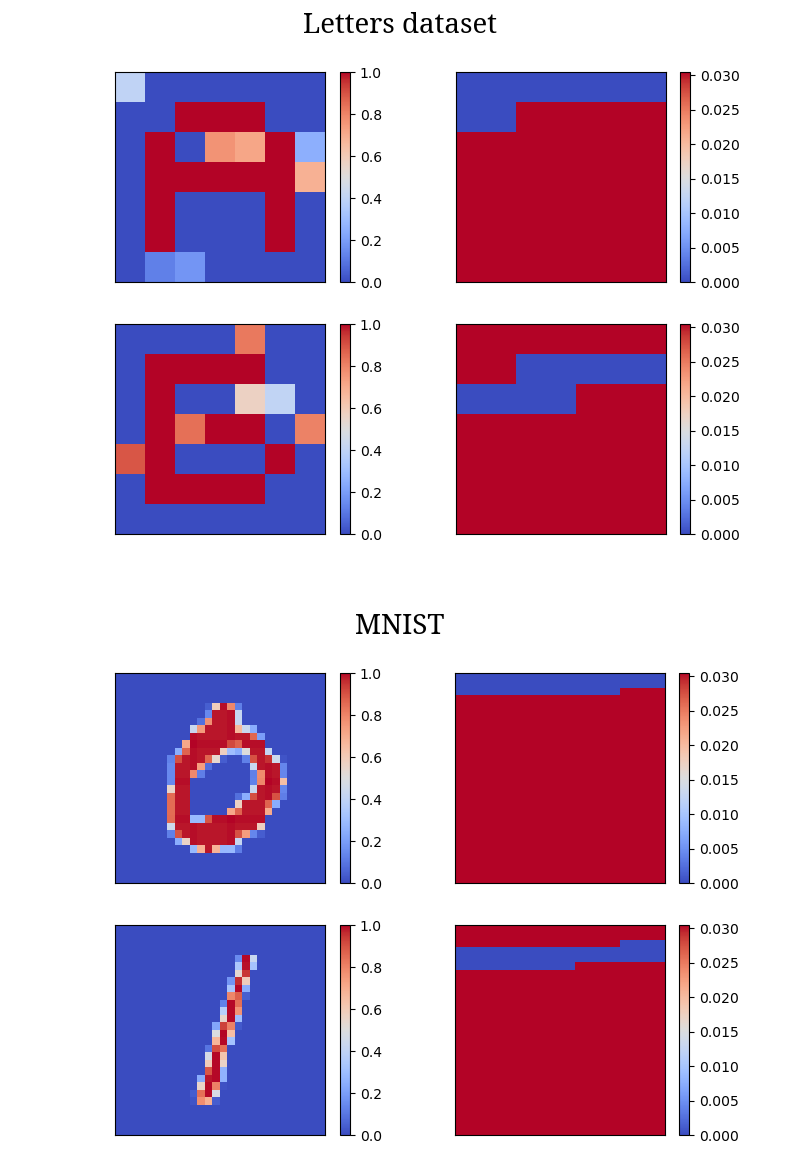}
    \caption{\footnotesize Examples of input from letters (left top panel) and  MNIST dataset (left bottom panel), shown alongside with their corresponding target attractors (on the right). In principle the asymptotic attractors can be assigned any patterns by employing the triplet $(0,x_m,x_p)$. In this case we have arbitrarily chosen to employ the null element and $x_p$, with no loss of generality.}
    \label{fig:examples_input_att}
\end{figure}

The vectors that function as attractors (and which can be depicted as images) are generated by arbitrarily employing just two of the three 
optional entries, specifically the null element and $x_p$. In Fig. (\ref{fig:examples_input_att}) a few representative items selected from their respective datasets and the corresponding crafted attractors are displayed to help intuition. We recall that our ultimate goal is  to associate each image (i.e., initial condition) with the corresponding final state (i.e., one of the fixed attractors), by appropriately adjusting the trainable parameters, as embedded in the spectral decomposition of the interaction matrix.\\

Operatively, we minimize the loss function $ \mathcal{L}=\frac{1}{|\mathcal{D}| }\sum_{j=1}^{|\mathcal{D}| } (\vec{x}_{j}-\vec{x}_{j}^{s})^\mathbf{T}(\vec{x}_{{j}}-\vec{x}_{j}^{s})$ where  
$\vec{x}$ is the value of $\vec{x}=\vec{x}(T)$, with $T$ large enough to allow for the dynamical system to approach its stationary state $\vec{x}^{s}$ (specific of the class to which the datum $j$, sampled from the train set $\mathcal{D}$ belongs to).  As anticipated, and in the minimal scheme here discussed, the evolution of the system is performed using Euler' algorithm, and the optimization is carried out by means of the Adam algorithm \cite{kingma2014adam}. In the following we will present the results obtained by employing the trained CVFR model as a classification algorithm against the two aforementioned datasets.

\begin{figure}
    \centering
    \includegraphics[width=0.45\textwidth]{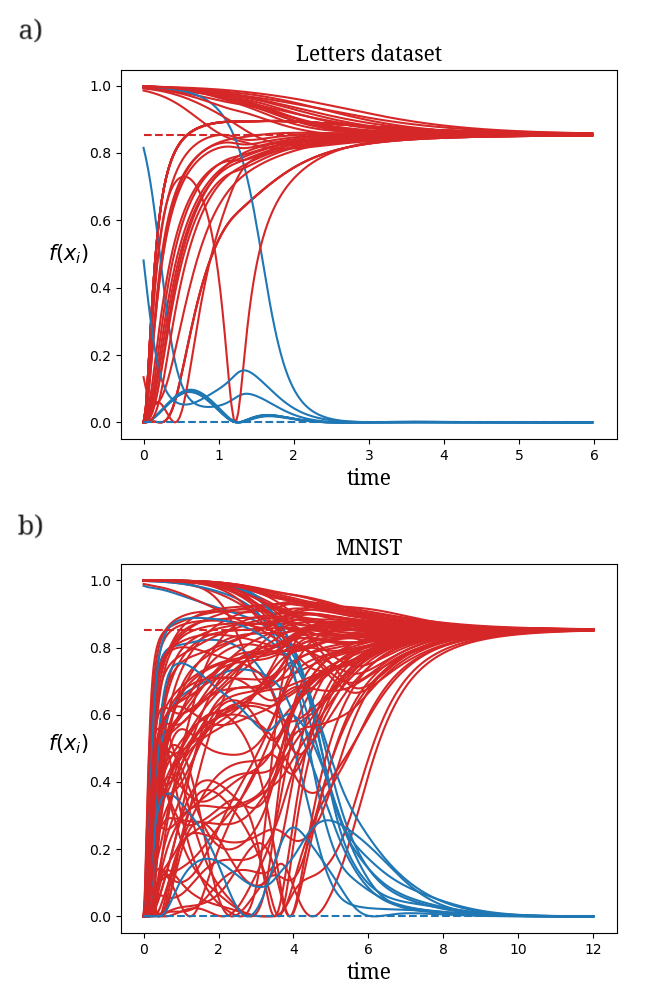}
    \caption{\footnotesize Temporal evolution of the neurons' activity (i.e., firing rate) in the trained network. Panel (a) refers to a data point from the letters dataset, while panel (b) shows the evolution of a data point from the MNIST dataset. The colors of the trajectories stand for the final state that the corresponding node must reach to achieve accurate classification. In both cases, the activity at $t=0$ reflects the intensity of the pixels in the image to be classified. After an initial time window displaying a seemingly irregular evolution, the trajectories converge to their respective target values, indicating that the model has correctly classified the input datum.}
    \label{fig:traj_det}
\end{figure}

To clarify the ensuing dynamic of the model upon training we refer to  
Fig. (\ref{fig:traj_det}) where the non linear image of state variable $\vec{x}$ via $f(\cdot)$ is plotted versus time. The choice of plotting 
$f(\vec{x})$ is simply reflecting graphical needs, and facilitates a better delivery of the message. Panel (a) refers to the letter dataset, while panel (b) is obtained by supplying as an input an image drawn from the  MNIST dataset. Colors assigned to the trajectories reflect the final value that each pixel (hence computing node) is expected to eventually approach for the system to reach the correct asymptotic attractor. Blue trajectories must converge to zero, while the red trajectories must converge to $f(x_p)$. As it can be visually appreciated, all depicted trajectories head towards the correct destination target (the red/blue solid curves cluster around the red/blue dashed horizontal lines), thus implying that, at late times, the system faultless stabilizes on the right attractor. The input is hence correctly identified and, as a consequence,  classified. Working in this setting, and as it should be clear from the above, classifying amounts to shaping the basin of attraction of the imposed stationary solution: items belonging to the same category generate an identical late-time pattern which therefore flags their association to the corresponding class. The stability of the imposed attractor, after training, can be assessed by diagonalizing matrix $\mathbf{J}$, as defined above. The eigenvalues are found to consistently populate the left portion of the complex plane (data not shown), thus implying that stability is a byproduct of the model's training.   

\indent To automatically quantify the accuracy of the algorithm, the following criterion was established. We calculate the inner product between the final state of the system $\vec{x}(T)$ and each of the 
imposed attractors $\vec{\psi}_k$, with $k=1,...,K$. Here  $T$ stands for the imposed time of integration, which should be sufficient for the system  to converge.  The initial data is then associated with the class corresponding to the attractor that maximizes the inner product, i.e., the attractor most aligned with the final state. This criterion is not overly conservative, as it allows to correctly classify data points even if their final states are not perfectly identical to the target attractor (either because they have not yet reached convergence or because they have fallen into spurious attractors similar to the correct one). Another possible criterion would be to assign the datum to its reference class  only if the $L^2$ distance between the final state and the target attractor is below an arbitrary threshold (resulting in a more conservative criterion). However, numerical analysis revealed that trajectories either converge quite decisively to the correct attractor or end up in a completely different state, and thus the two above mentioned  criteria yield substantially similar results. In particular, for the letters dataset the deterministic trained CVFR model reaches an accuracy on the test set of $99.9\%$, while for the MNIST dataset  the measured accuracy is $97.2\%$.

\section{On the stochastic version of the CVFR model: opposing adversarial attacks}
\label{sec:stochastic formulation}

Noise, be it endogenous or exogenous, can occasionally - and unintuitively -  yields beneficial effects in several contexts of applied and fundamental relevance. Motivated by this general understanding, we here consider an extended version of the CVFR which includes a stochastic contribution and elaborate on the role played by the additional noisy component, when performing the assigned classification task.

More specifically, we shall consider a multiplicative noise term that shakes the deterministic formulation of the CVFR model as specified by eq. 
(\ref{eq:model_definition}). The amplitude of the noise is a function of the state variable and it is designed so as to fade away when the system eventually reaches any of the planted asymptotic attractors.The reason for this choice is that it allows the same (hence deterministic) loss function to be employed when training the stochastic version of the model. The imposed noise will solely act out-of-equilibrium to provide the dynamical classification algorithm with an additional degree of flexibility, in the search for the correct destination target. 

The Langevin equation that defines the evolution of the noisy system is:
\begin{equation}
    \label{eq:Langevin_equation}
    \dot x_i =-x_i +\frac{1}{\sqrt{N}} \sum_j^N A_{ij}f(x_j) + \eta_i(t)d(\vec{x}(t)),
\end{equation}
where $\eta_i(t)$ is a delta correlated Gaussian random variable with zero mean and standard deviation $\sigma$. The amplitude factor $d(\vec{x}(t))$ reads: 

\begin{equation}
    d(\vec{x}(t)) = \tanh \bigg (\sqrt[K]{\prod_{k=1}^{K}\frac{||\vec{x}(t)-\vec{x}_{k}^s||^2}{N}} \bigg ).
    \label{eq:damping_factor}
\end{equation}

and it is constructed so as to progressively dampen the noise term when the planted attractors are approached. In other words, the deterministic model is eventually recovered when getting close to the aforementioned attractors. In this limit the analysis developed in the preceding Section holds valid. As we will show in the following, working with the stochastic version of the model, yields trained solutions that are more robust against random perturbations of the input. The stochasticity acts as an internal regularizer, in line with previous observations \cite{benedetti2023training, liu2019neural}. 

To quantify the impact of the imposed multiplicative noise $\vec{\eta}$ we set a value of the amplitude $\sigma$ and train the stochastic version of the model, as specified by Eq. (\ref{eq:Langevin_equation}), against the considered datasets. The average computed accuracy is respectively reported in  Tables \ref{table1} and \ref{table2}. The reported values were obtained by repeating the experiments five times. The results appear to be very robust, as the standard deviation for the accuracy is less than 0.1\% for all tested settings. Hence, only the significant figures are reported in the tables. For the letters dataset the deterministic setting ($\sigma=0.0$) and one of the stochastic models ($\sigma=0.05$) yield the best accuracy, but the gap with the other models in terms of recorded performance is tiny. When it comes to the MNIST dataset,  three out of the four employed stochastic models (ordered for ascending values of imposed $\sigma$ in Table \ref{table2}) display slightly better accuracy than that reported for the  corresponding deterministic model. However, all measured accuracy are indeed in line to typical values reported for the celebrated MNIST tested reference. 

\begin{table}
    \begin{tabular}{ |p{1.5cm}|p{1cm}|p{1cm}|p{1cm}|p{1cm}|  }
 \hline
 \multicolumn{5}{|c|}{Letters dataset }\\
  \hline
$\sigma$ & \textbf{0.0}  &	\textbf{0.05} &	0.1 &	0.2\\
 \hline
accuracy &\textbf{99.9\%} &	\textbf{99.9}\% &	99.8\% &	99.8 \%\\
 \hline
\end{tabular}
\caption{Accuracy on the \textit{letters} dataset for different levels of inherent noise in the model.}
\label{table1}
\end{table}

\begin{table}
    \begin{tabular}{ |p{1.5cm}|p{1.2cm}|p{1.2cm}|p{1.2cm}|p{1.2cm}|p{1.2cm}|}
 \hline
 \multicolumn{6}{|c|}{MNIST}\\
  \hline
$\sigma$ & 0.0 &	\textbf{0.1} &	0.5 &	0.8 & 1.0\\
 \hline
accuracy &97.2\% &	\textbf{98.0\%} &97.8\% &	97.6\% & 97.2 \%\\
\hline
\end{tabular}
 \caption{Accuracy on the \textit{MNIST} dataset for different levels of inherent noise in the model.}
\label{table2}
\end{table}

\begin{figure}
    \centering
    \includegraphics[width=0.45\textwidth]{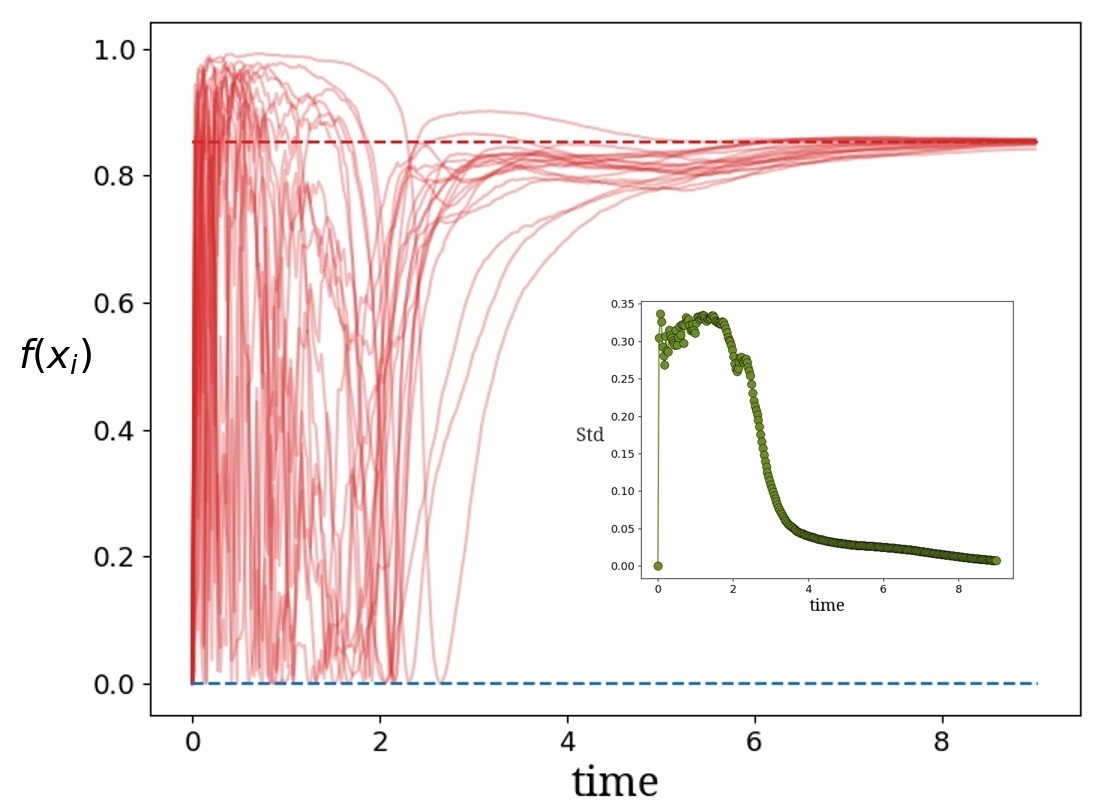}
    \caption{\footnotesize Multiple temporal evolutions of non linear image $f(x_i)$ of the state variable $x_i$, for a fixed node $i$, and  for a given data input selected from the MNIST dataset. The depicted curves refer to different realizations of the trained stochastic model with $\sigma=1.0$. The variability of the trajectories depends solely on the stochastic nature of the - post-trained - evaluation process. As time progresses, all trajectories converge to the target fixed point, and the stochasticity fades away, as follows the damping factor
    (\ref{eq:damping_factor}). The inset shows the standard deviation of the generated trajectories, as computed numerically: at $t=0$, there is no variability (the initial data point is always the same), while at subsequent times, the standard deviation increases, before eventually converging to zero at late times.}
    \label{fig:same_pixel}
\end{figure}

\indent Furthermore, it is instructive to analyse the dynamics of the trained model, by just focusing on the temporal evolution of the signal, as displayed by one representative node of the collection. To this end, we generate multiple trajectories of the trained model, subject to the very same input data. The stochastic differential equation (\ref{eq:Langevin_equation}) is numerically integrated, upon training, by resorting to the classical Euler-Maruyama algorithm.  The inherent variability of the recorded trajectories solely stems from the stochastic nature of the model, as both input data and the selected node stay unchanged across distinct realizations of the examined dynamics. In Fig. \ref{fig:same_pixel}, the time evolution of state variable $x_i$, filtered via the non linear function $f(\cdot)$, is reported, for a stochastic model trained to classify the MNIST images, with noise amplitude set to $\sigma=1.0$. Individual trajectories exhibit a significant degree of variability, during an initial stages of the evolution and before reaching the target equilibrium. The model parameters are adjusted, during the training, in such a way that inputs  
belonging to a specific class converge towards their corresponding fixed attractor for (almost) every realization of noise that is supposed to shake the underlying deterministic flow. The variability displayed by the recorded trajectories can be measured over time by calculating the standard deviation of the activities (the non linear transformation of the state variable $x_i$), as measured at different time points. This quantity is reported in the inset of Fig. \ref{fig:same_pixel}. At $t=0$, all trajectories originate from the very same initial condition (the input data is kept fixed) and the associated standard deviation is hence zero, by definition. Then, at $t \ne 0$, when the system explores its out of equilibrium landscape to find the path towards equilibrium,  large standard deviations are reported to occur, an obvious proxy of the inherent variability as displayed by the stochastic classifier. Eventually, the standard deviation drops to zero as the damping factor (\ref{eq:damping_factor}) turns the stochastic model into its deterministic analogue. The absence of noise at late times prevents the trajectories from diverging again.

\begin{figure}
    \centering
    \includegraphics[width=0.45\textwidth]{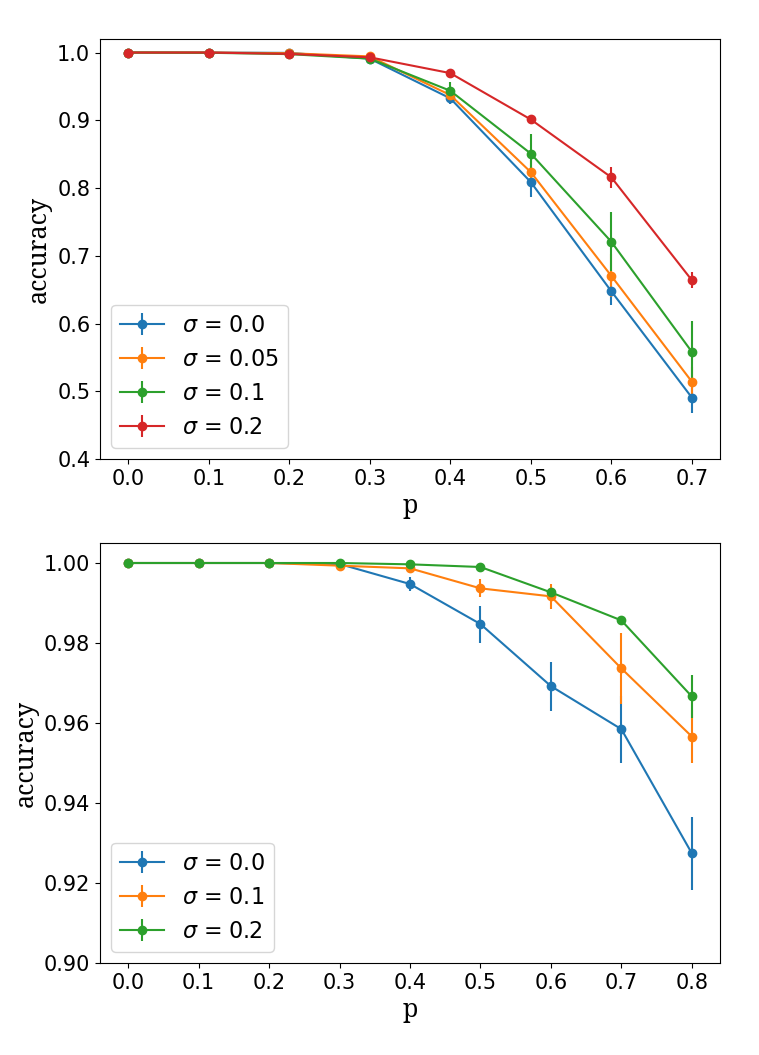}
    \caption{\footnotesize \textbf{Letters dataset}. The accuracy on the test set is reported as a function of the parameter $p$ that quantifies the intensity of the attack. The upper panel refers to a type A attack, while the lower panel refers to a random type B attack (see the main body of the manuscript). Different lines correspond to different models. Each model was trained with a different level of inherent noise, as quantified by the amplitude parameter $\sigma$,  shown in the legend. Specifically, $\sigma = 0.0$ corresponds to the deterministic case. As expected, the models become progressively less effective as the parameter $p$ increases. However, the stochastic models show greater resistance to external noisy attacks as compared to their deterministic analogues.}
    \label{fig:robustness_MULTIPLOT_letters}
\end{figure}

\begin{figure}
    \centering
    \includegraphics[width=0.45\textwidth]{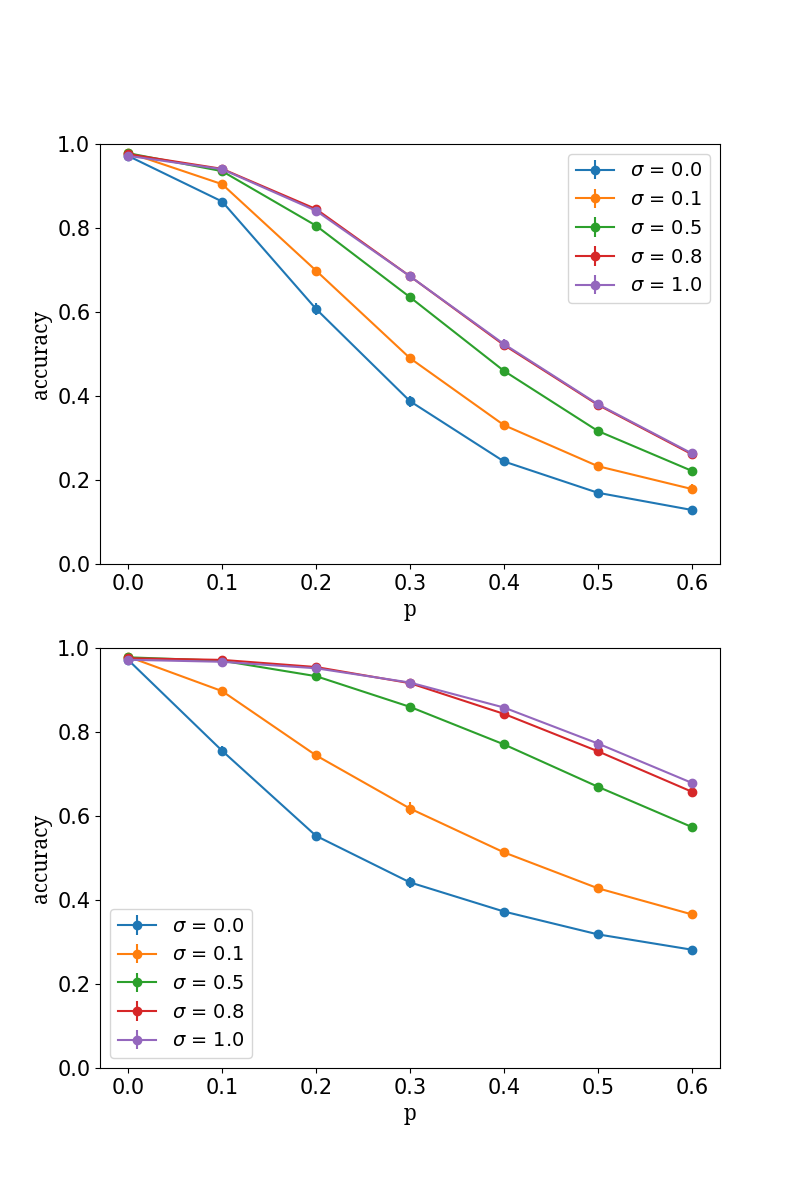}
    \caption{\footnotesize \textbf{MNIST dataset}. Accuracy calculated on the test set as a function of the parameter $p$ that quantifies the intensity of the attack. The upper panel refers to a type A attack, while the lower panel refers to a random type B attack (see the main body of the manuscript). Different lines correspond to different models. Each model was trained with a different level of noise, as quantified by the amplitude parameter $\sigma$, see legend. Specifically, $\sigma = 0.0$ corresponds to the deterministic case. We note that as the parameter $p$ increases, all models become progressively less effective. However, the stochastic models show greater resistance to noisy attacks as compared to the corresponding deterministic setting.}
    \label{fig:robustness_MULTIPLOT_mnist}
\end{figure}

To elaborate further on the role played by noise, and demonstrate that stochasticity offers a clear operational advantage, we consider the response of the trained model when subject to different types of adversarial attacks. Specifically, we tested two different types of attacks, both dependent on a parameter $p$ that controls the intensity of the imposed noise disturbance. The first type of employed attack (A) follows the procedure adopted to generate the database of noisy letters. It  consists in replacing the true value as displayed by the supplied item with random values drawn from a uniform distribution between 0 and 1, for a percentage equal to $p$ of the image's pixels. The second type of attack (B) assumes that a random value, drawn from a uniform distribution spanning the interval $-p$ to $p$, is simultaneously added to all the image's pixels. Notice that, both for case A and B, the limiting condition $p=0$ corresponds to the absence of adversarial noise, while increasing values of $p$ signifies higher levels of image corruption. The performances of the trained models (either in the stochastic or deterministic versions) are tested against the corrupted datasets. The obtained accuracies as a function of the parameter $p$ are shown in Fig. \ref{fig:robustness_MULTIPLOT_letters} for the letters dataset and in Fig. \ref{fig:robustness_MULTIPLOT_mnist} for the MNIST dataset.  Stochastic models are more resistant to the attacks, as compared their deterministic counterparts: indeed, they yields remarkably high performance as the level $p$ of imposed corruption increases.   The higher the inherent stochasticity, as quantified by the amplitude parameter $\sigma$, the more resistant the models are to the external attacks, within the range of explored parameters. Further increasing the noise level leads to excessive randomness that prevents the model from converging during  training. The maximum values of $\sigma$ for a swift training convergence  are $\sigma=0.2$, for the letters dataset, and $\sigma=1.0$ for MNIST. These are therefore the optimal values for having robust stochastic models, to oppose external random attacks of the type here investigated.

\section{Conclusion}
\label{sec:conclusion}

In this work we have studied a variant of the SA-nODE algorithm for classification, which is based on the celebrated \textit{Continuous-Variable Firing Rate} (CVFR) model \cite{kim2019simple}. This is a reference scheme employed in computational neuroscience to account for the interlaced dynamics of biological neurons. By exploiting a spectral decomposition of the coupling matrix, we achieve the goal of planting into the model a family of non linear attractors, by using a limited alphabet of digits that follows analytically from the stationarity conditions of the examined problem. The CVFR model, also known in the literature as continuous Hopfield model \cite{movellan1991contrastive}, is then trained as a classification algorithm, by forcing items belonging to distinct classes, and supplied as initial conditions to the dynamical system, to head towards the corresponding assigned attractor. As in the spirit of the SA-nODE procedure, learning (to classify) amounts to sculpt the basin of attractions of a veritable - continuous - dynamical model that, in the case here examined, bears interest for  neuroscience applications. Further, in the second part of the paper, we extended the SA-nODE recipe to account for a stochastic version of the CVFR model. While the performance of  deterministic and stochastic models are in line, at least for the limited class of datasets here explored, the stochastic setting proves definitely more efficient in opposing random adversarial attacks that perturb the supplied input data. This is yet another of the very many surprising and, to some extent, un-intuitive phenomena to be ascribed to noise, in its diverse and multifaceted manifestations.

\begin{acknowledgments}
The work of L.C., D. Fanelli, D. Febbe and R. M. is supported by \#NEXTGENERATIONEU (NGEU) and funded by the Ministry of University and Research (MUR), National Recovery and Resilience Plan (NRRP), project MNESYS (PE0000006) "A Multiscale integrated approach to the study of the nervous system in health and disease" (DR. 1553 11.10.2022).\\
F.D.P. acknowledges the funding from Project ‘Mathematical modelling for a sustainable circular economy in ecosystems’ (grant n. P2022PSMT7) funded by EU in NextGenerationEU plan through the Italian ‘Bando Prin 2022 - D.D. 1409 del 14-09-2022’ by MUR.\\
F.D.P. thanks Gruppo Nazionale di Fisica Matematica of Istituto Nazionale di Alta Matematica for partial financial support.\\
F.D.P. acknowledges the funding from NextGenerationEU PRIN2022 grant no. 2022P5R22A “The Mathematics and Mechanics of Nonlinear Wave Propagation in Solids”.

\end{acknowledgments}





\bibliography{apssamp}

\end{document}